\documentclass[conference, 12pt]{IEEEtran}

\usepackage[T1]{fontenc}
\usepackage[utf8]{inputenc}
\usepackage{lmodern}
\usepackage[colorlinks=true,urlcolor=cyan]{hyperref}

\usepackage{xspace}
\usepackage{authblk}
\usepackage{cite}
\usepackage{amsmath,amssymb,amsfonts}
\usepackage{algorithmic}
\usepackage{graphicx}
\usepackage{textcomp}
\usepackage{subcaption}
\usepackage{xcolor}
\usepackage{listings}

\usepackage{tikz}
\usepackage{pgfplots}
\usepackage{pgfplotstable}
\usetikzlibrary{matrix,positioning,arrows,fit,patterns, shapes, shadows.blur}
\usetikzlibrary{arrows, arrows.meta}

\colorlet{punct}{red!60!black}
\definecolor{background}{HTML}{EEEEEE}
\definecolor{delim}{RGB}{20,105,176}
\colorlet{numb}{magenta!60!black}

\lstdefinelanguage{json}{
    basicstyle=\normalfont\ttfamily,
    frame=lines,
    backgroundcolor=\color{background},
    literate=
      {:}{{{\color{punct}{:}}}}{1}
      {,}{{{\color{punct}{,}}}}{1}
      {\{}{{{\color{delim}{\{}}}}{1}
      {\}}{{{\color{delim}{\}}}}}{1}
      {[}{{{\color{delim}{[}}}}{1}
      {]}{{{\color{delim}{]}}}}{1},
}

\newcommand{\blenderName}{BlenderProc\xspace}

\newcommand{\cnns}{convolutional neural networks\xspace}
\newcommand{\refSec}[1]{section \ref{#1}\xspace}
\newcommand{\refFig}[1]{figure \ref{#1}\xspace}
\newcommand{\key}[1]{"#1"}

\newcommand{\etal}{et al.\xspace}

\newcommand{\set}[2]{\pgfmathsetmacro{#1}{#2}}

\title{\blenderName\\{\large \textbf{\url{https://github.com/DLR-RM/BlenderProc}}}}
\author[1]{Maximilian Denninger}
\author[1]{Martin Sundermeyer}
\author[1]{\\Dominik Winkelbauer}
\author[1]{Youssef Zidan}
\author[1]{Dmitry Olefir}
\author[1]{\\Mohamad Elbadrawy}
\author[1]{Ahsan Lodhi}
\author[1]{Harinandan Katam}
\affil[1]{German Aerospace Center (DLR)}
\begin{document}
\makeatletter
\let\@oldmaketitle\@maketitle
\renewcommand{\@maketitle}{\@oldmaketitle
\center
\newcommand{\imageWidthEyeCatcher}{0.65\columnwidth}
\includegraphics[width=\imageWidthEyeCatcher]{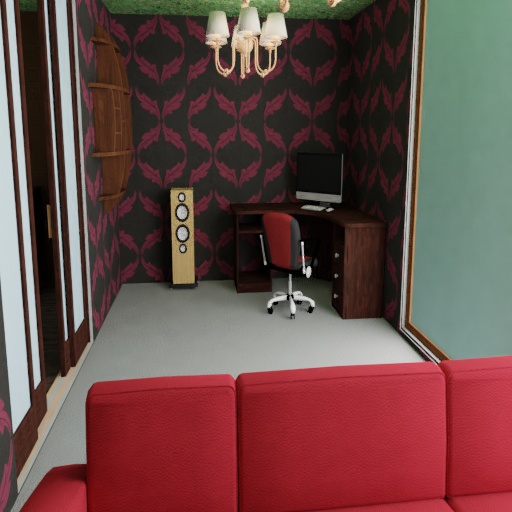}\quad\includegraphics[width=\imageWidthEyeCatcher]{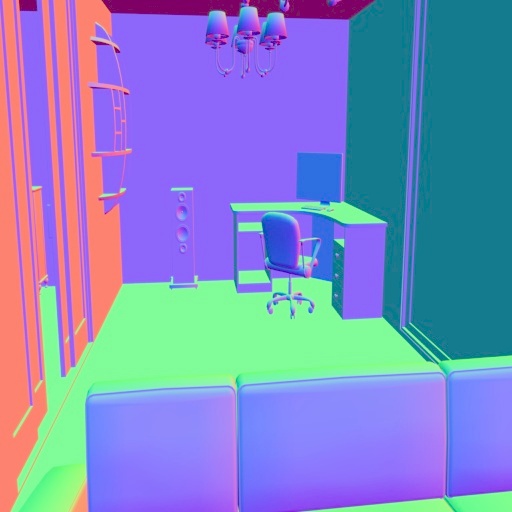}\\ \vspace{10pt}\includegraphics[width=\imageWidthEyeCatcher]{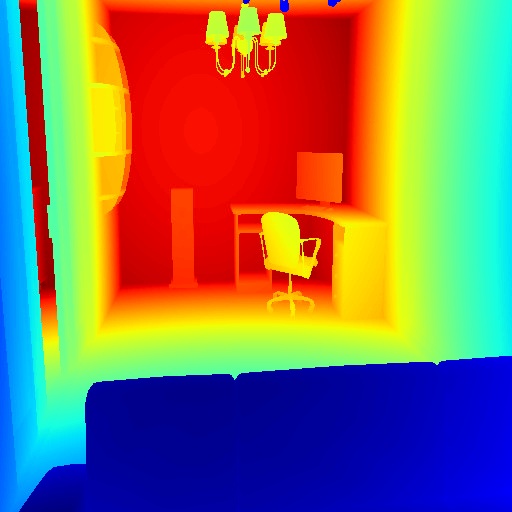}\quad\includegraphics[width=\imageWidthEyeCatcher]{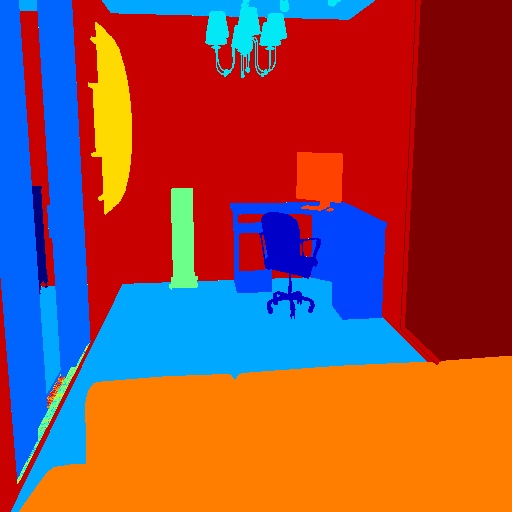}\captionof{figure}{The \blenderName can generate color, normal and depth images and also class segmentation and instance segmentation masks, which can be used for the training of \cnns.}
\label{fig:eye-catcher}
}
\makeatother

\maketitle

\begin{abstract}
\blenderName is a modular procedural pipeline, which helps in generating real looking images for the training of \cnns.
These can be used in a variety of use cases including segmentation, depth, normal and pose estimation and many others.
A key feature of our extension of blender is the simple to use modular pipeline, which was designed to be easily extendable.
By offering standard modules, which cover a variety of scenarios, we provide a starting point on which new modules can be created.
\end{abstract}

\section{Introduction}

In the past years the demand for high quality images, which can be used for training has risen drastically.
In order to generate these images in certain domains like pose estimation or instance segmentation, a horde of students has to be used to single-handedly mask and label all the images.
This is expensive and time consuming, to avoid this we created \blenderName. 

\blenderName is a modular procedural pipeline, which can generate real looking images of scenes, which provide perfect segmentation masks, depth or normal images.
For that we use the open source project Blender\cite{blender}, which provides a nice and clean python API, on which our pipeline was built. 

This enables us to generate perfect training data even in scenarios where this might be challenging.
We use \blenderName so far with the SUNCG dataset \cite{suncg}, all objects used in the bob challenge \cite{hodan2018bop} and we enable normals and depth reconstruction in the replica dataset \cite{straub2019replica}. 

Of course we are deeply interested in increasing the replacement of standard OpenGL Renderer in as many datasets possible. 
That's why we make the source code open source and we will keep contributing to it and are hoping of the contributions of others.

A main objective in the design of our pipeline is keeping it easy to use, simple to extend and clear to read.
We provide an in-depth documentation to all functions and modules, which include their usage and the possible settings values to configure each module to the problem at hand. 
Integrating new modules, including renderer for normal or depth images, can be done by just adding the name of the corresponding module to the settings file.

In comparison to an OpenGL implementation speed was not our main objective, as we see data as something, which is less often changed as an architecture.
So spending one or two days on generating the necessary amount of data, is an acceptable time demand for us.
Of course, this makes it hard to use in scenarios, where videos or reinforcement learning loops are used.
Nonetheless, we are committed to improve the speed of the data generation.

\section{Related work}
\label{sec:related_work}

In various computer vision tasks the use of synthetic images for benchmarking and training of \cnns is widely done.

Most approaches train on images generated with OpenGL pipelines based on simple rasterization techniques.
Su \etal \cite{Su2015} estimated the viewpoint by generating images of 3D object models.
Hinterstoisser \etal \cite{Hinterstoisser2018} synthesized images for object instance segmentation  and Dosovitskiy \etal \cite{Dosovitskiy2015} for optical flow estimation. Sundermeyer \etal \cite{Sundermeyer2019} computed images for 6D pose estimation, where they randomized the parameters to generate objects and pasted those in front of random selected real photographs.

All of those approaches are easy to implement however, the resulting images are not realistic.
The shading of the objects is not consistent and similar to the real world, which plays an important role if several objects are in close proximity to each other.
These missing interactions between each other are completely neglected in such pipelines, where the main asset is speed. 
Furthermore, sampling in these environments happens usually randomly that not even physical constraints are applied, which makes the scenes even more unrealistic.
In particular in the instances were photos are copied in front of real images the shading is complete inconsistent.
For example Movshovitz-Attias \etal \cite{Movshovitz-Attias2016} showed the benefit when using photo realistic images of 3D car models placed within real looking 3D scene models over naive rendering methods.

Li and Snavely \cite{Li2018} train models for intrinsic image decomposition with Physically based rendering (PBR) images.
Zhang \etal \cite{Zhang2017} use these images for semantic segmentation, normal estimation and boundary detection, but where they use Mitsuba \cite{mitsuba}, we use blender-integrated cycles and get nicer looking images in less time. 
Other ways of creating PBR images haven been also proposed \cite{Wood2015, Wood2016, Shrivastava2017}.
Hoda\~n \etal \cite{Hodan2019} also presented a nearly perfect PBR simulation in his work about object instance detection. However, they are not using a open source backend, which means even when they release their source code, people would have to buy the required license of Autodesk Maya.

As just proven we are not the first to do PBR, but our main objective is a different one in contrast to prior work.
Our main goal is designing a pipeline, which can be used by any one in Computer Vision field to generate the data they need by providing a good starting point and even complete solutions for a lot of datasets.
By providing a good starting point and even complete solutions for a lot of datasets.

\section{Modules}
\label{sec:modules}

As mentioned earlier our pipeline consists of several modules, which can be easily configured over a config file, which is explained more in depth in \refSec{sec:config}.
These modules can change, for example, the world state in blender by loading new objects or sampling new camera positions.
Not all modules do that though, some, like the color renderer (see \refSec{sec:color_renderer}), only run through all key points of the cameras and render the given scene.
Each pipeline run includes the definition of the camera positions. 
These positions are saved as key points for a camera object. 
A key point is the location and rotation of a certain object at a certain time in the animation run, which we use to render several images per scene.
Each key point then generates a corresponding image.
After the loading of the camera positions the scene is loaded (see \refSec{sec:loader}), this includes some objects and light sources.
The final step is the desired rendering, the color renderer and the depth renderer might be executed after each other.
The final result will be collected and stored in an hdf5 file\cite{hdf5}. 
The keys of it can be specified in the config file.
This process enables it to easily generate images, where the labels and all corresponding data is stored in the same file, which is also compressed.

\section{Config}
\label{sec:config}

As general configuration we use YAML files, which we parse for certain specific keys, we introduce in this chapter.

Each config file contains three parts, the first is the setup: 

{
\small
\begin{lstlisting}[language=json]
"setup": {
  "blender_install_path": "/PATH",
  "blender_version": "blender-2.80",
  "pip": [ 
    "h5py", 
    "imageio"
  ]
}
\end{lstlisting}
}
It contains the blender install path, where blender will be installed, if you don't have it already. 
We advise installing blender again for the tests to avoid that the installation of new packages in the blender python distribution will cause any problems for other simulations you might run. 
The key \key{pip} stores all packages, which will be additionally installed in the blender python distribution, here you can add the packages, you need for your own modules. 
Important to note is that blender comes with its own python distribution, which is independent of your system wide installation.

However, the run.py script will be executed by your own python distribution and has some requirements, which will be checked when it is first executed.

The second part named \key{global} describes settings, which are available in all modules. In contrast to that are all the settings, which are specified in the \key{modules}, those are only accessible in the corresponding module. 
The \key{output\textunderscore dir} is set here to \key{<args:0>}, this means, the run.py takes additional parameters, which then get feed into these values. Here \key{<args:0>} stands for the first value after the config file path.
This makes it easy to pass values from the command line to the relevant module.
{\small
\begin{lstlisting}[language=json]
"global": {
  "all": {
    "output_dir": "<args:0>"
  }
}
\end{lstlisting}}
The last part are the \key{modules}, where the order in the config determines the execution order. 
Each module has a name, which corresponds to the name of the module folder and the class name, the pipeline will load the corresponding class dynamically.
This class has to be derived from our module base class and will execute the code inside of it, the corresponding settings here will be passed to it.
{\small
\begin{lstlisting}[language=json]
"modules": [
  {
    "name": "main.Initializer"
  },
  {
    "name": "loader.ObjLoader",
    "config": { "path": "<args:1>" }
  }
]    
\end{lstlisting}
}
In this example the \key{main.Initializer} sets up the blender environment at first and afterwards the \key{loader.ObjLoader} is run. 
This module can load a single .obj file into the scene.

Each module has at the top of its class definition an explanation of all the config parameters, which can be set for this module.

\section{Loader}
\label{sec:loader}

The loader modules load all kinds of 3D meshes into the scene and furthermore all other objects like lamps and cameras.
The basic loader as shown in the last section can just load one .obj file.
We also provide a loader for the suncg-scenes renders, which were shown in \refFig{fig:eye-catcher}.
This loader loads all the .obj files and the corresponding textures and categories, which we later use in the segmentation renderer (see \refSec{sub:seg-renderer}).
Additionally it loads all camera poses, to do that we generate a generic camera loader, where you can specify the structure of your camera positions files, which will then be loaded into blender.
If you want to load more than a few objects and want to do a semantic segmentation, you will most likely have to write your own loader.
This generic camera loader style can also be used to load lights.

\section{Renderer}
\label{sec:renderer}

A renderer in this work is anything that produces an image like output of a given scene.
It is tightly coupled to the camera key points, which positions are defined in the loader.
The standard one is the color renderer, which generates an rgb image for each camera pose of the given scene.

Each renderer has several options, a few are explained here:
\begin{itemize}
\item \key{resolution\_x, resolution\_y}: The resolution of the images to render
\item \key{samples}: The amount of samples used to render a scene, more samples reduce the noise
\item \key{render\_depth}: Renders the depth of the current image (see \refSec{sub:depth-renderer})
\item \key{stereo}: Enables the rendering of stereo images
\end{itemize}

\subsection{Color renderer}
\label{sec:color_renderer}

\begin{figure}[h]
\center
\includegraphics[width=0.6\columnwidth]{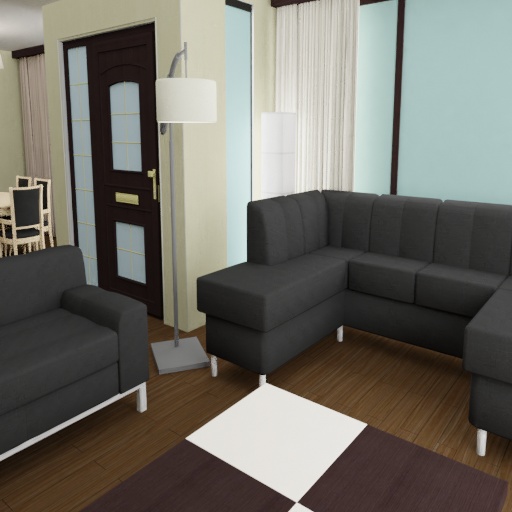}\caption{Rendered color image of a suncg scene\cite{suncg}}
\label{fig:color-render-image}
\end{figure}

The color renderer produces realistic looking images of a given scene, as depicted in \refFig{fig:color-render-image}.
This image was generated based on the data from the suncg dataset\cite{suncg}.
In order to get enough light in these scenes, we made the ceiling emit light, because not all rooms in this dataset contained lamps and the placement of them did not lead to a sufficient and realistic illumination.

The color renderer has additional options, some are explained here:
\begin{itemize}
\item \key{min\_bounces, max\_bounces}: Restricts the number of bounces each ray can do; the higher, the better the quality of the results, but the lower the rendering speed
\item \key{glossy\_bounces}: Maximum number of glossy reflection bounces, has to be bigger than zero, such that glossy effects are visible
\end{itemize}

\subsection{Depth renderer}
\label{sub:depth-renderer}

\begin{figure}[h]
\center
\includegraphics[width=0.6\columnwidth]{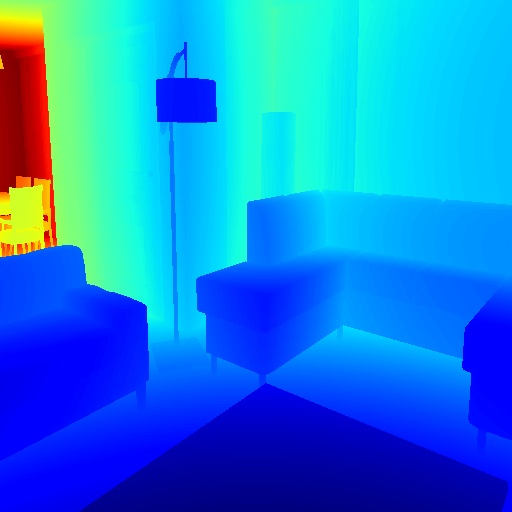}\caption{Depth image of a suncg scene\cite{suncg}}
\label{fig:depth-render-image}
\end{figure}

This renderer generates depth images based on the internal depth estimation of blender.
This might mean that the results do not have a perfect precision based on the fact that the Z-Buffer is used to generate the results.
So far we did not evaluate the precision of this. 
Furthermore, it is to note that the depth renderer can only be used with a different renderer in combination, because the depth rendering is a byproduct of another renderer.
The settings flag \key{render\_depth} should than be set to true.
Furthermore, the key \key{depth\_output\_key} can be changed to set the key in the .hdf5 file.

In the case that not the whole image is covered by objects all empty pixel will be set to infinity.

\subsection{Normal renderer}
\label{sub:normal-renderer}

\begin{figure}[h]
\center
\includegraphics[width=0.6\columnwidth]{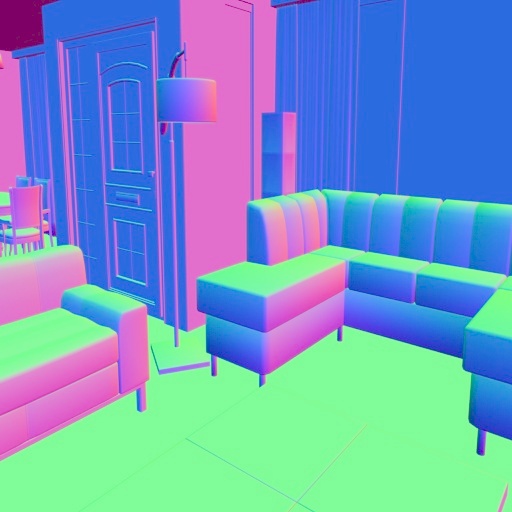}\caption{Normal image of a suncg scene\cite{suncg}}
\label{fig:normal-render-image}
\end{figure}

Many deep learning tasks greatly improve by introducing normals as an input to the training problem. 
One of those problem is for example depth estimation.
The normals are normalized and presented as XYZ values.

\subsection{Segmentation renderer}
\label{sub:seg-renderer}

\begin{figure}[h]
\center
\includegraphics[width=0.6\columnwidth]{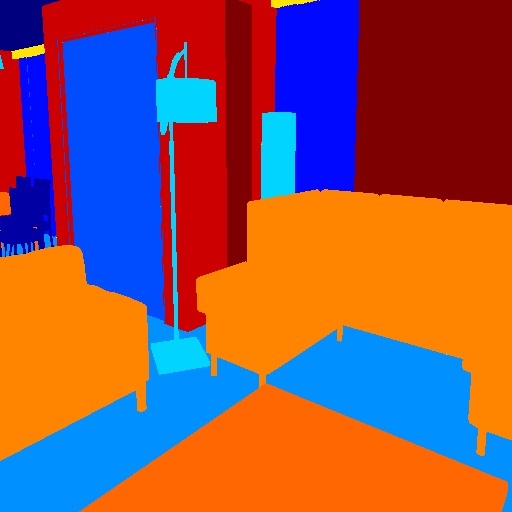}\caption{Segmentation mask of a suncg scene\cite{suncg}}
\label{fig:seg-render-image}
\end{figure}

For the segmentation to work each object needs a category id, in the suncg loader they are already set correctly. 
For your own loader, you would have to do this manually, by adding the attribute \key{category\_id} to all objects.
This attribute contains the number of the class in your the dataset.

The output is saved as an one dimensional 2D image, where the classes are stored as values from 0 to the number of classes specified.
It is also possible to create instance segmentation masks, where not only the indices are saved but also a dictionary, which maps the indices to a class and an instances number.
Both of them are stored in the .hdf5 file, which makes the usage easier.

To switch between semantic and instance segmentation the following setting is available:
\begin{itemize}
\item \key{map\_by}: If set to \key{class}, then semantic segmentation based on the \key{category\_id} is done. If set to \key{instance}, then instance segmentation is performed
\end{itemize}

\section{Sampler}

\blenderName contains several samplers, for lights, cameras and objects in general.
All of those can be specified over the settings, this process was designed to be as simple as possible.
We already provide a variety of samplers to choose from.
We have basic samplers like sampling in a bounding box or a sphere and more complex ones where you can sample on the shell of a sphere or collision free in a bounding box.

Here are two examples of how to easily use samplers when generating light sources or camera locations.
{
\small
\begin{lstlisting}[language=json]
"location": {
  "name": "Uniform3dSampler",
  "parameters": {
    "min": [0, 0, 0],
    "max": [1, 1, 1]
  }
}
\end{lstlisting}
}
\begin{figure}
{
\small
\begin{lstlisting}[language=json]
"location": {
  "name": "SphereSampler",
  "parameters": {
    "center": [0, 0, 1],
    "radius": 4,
    "mode": "SURFACE"
  }
}
\end{lstlisting}
}
\end{figure}
Furthermore, each sampler can perform proximity checks.
This means that for example a camera might be sampled and then checked if the position is at least a meter away from a any object and also that the mean distance of objects in the scene is in a certain range.
Those options can be arbitrarily logically combined, which makes creating the perfect conditions easy.
In this example it is required that a camera is at least one unit away from any object in its view and that the average distance over its values is between one and four unit. Only  then is the pose used.
{
\small
\begin{lstlisting}[language=json]
"proximity_checks": {
  "min": 1.0,
  "avg": {
    "min": 1.0,
    "max": 4.0
  }
}
\end{lstlisting}
}

\section{Generation Time}

The generation is done offline in a batch fashion, if the camera poses are already sampled then only the specific renderers are called per scene and in the end all the files per camera pose are combined to an .hdf5 file.
For the suncg files we could generate around 3.000 images per hour on a single gpu. 
However, these are rather complex scenes and in the most scenarios the rendering should even be faster than this.
Nonetheless, was speed no major concern for us, as we think generating is usually not a repeating process. 
So spending a weekend on it and then having 180.000 images is usually enough for most training tasks.
In general the generation of anything except the color is quite fast, because only for those cases multi-sampling is necessary.

Currently, the generation time is mostly slowed by rebuilding a spatial tree of the scene for every frame, this is currently a missing feature, where we are already in contact with the blender foundation if we can get support for this.

\section{Examples}
\label{sec:examples}

This chapters contain more examples images from the \blenderName.
In \refFig{fig:example_imgs} ten scenes are shown with the corresponding color, normal, depth and segmentation images. 
Each set would be saved in one compressed .hdf5 container file.

\newcommand{\plotAll}[4]{
\begin{tikzpicture}

\set{\firstNr}{int(#2)};
\set{\correctNr}{int(\firstNr + #4)};
\set{\columnFac}{#3};
\newcommand{\imageWidth}{\columnFac\columnwidth};
\set{\fac}{1.05};
\node at (0,0) {\includegraphics[width=\imageWidth]{#1_colors_\firstNr.jpg}};\node at (\imageWidth*\fac,0) {\includegraphics[width=\imageWidth]{#1_normals_\firstNr.jpg}};\node at (0,\imageWidth*-\fac) {\includegraphics[width=\imageWidth]{#1_depth_\correctNr.jpg}};\node at (\imageWidth*\fac,\imageWidth*-\fac) {\includegraphics[width=\imageWidth]{#1_seg_\correctNr.jpg}};
\end{tikzpicture}
}
\set{\imgSize}{0.2175}
\begin{figure}
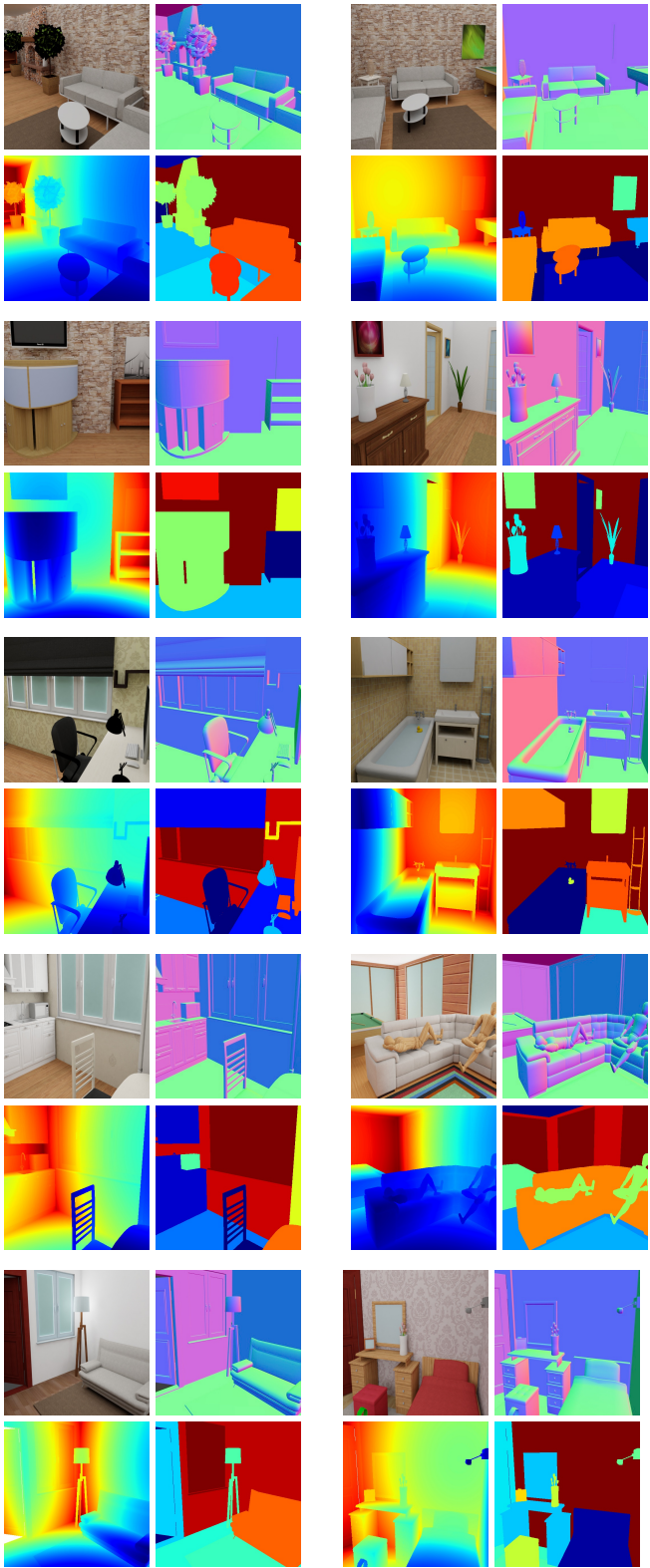

\plotAll{figures_first_room}{0}{\imgSize}{1}\plotAll{figures_first_room}{1}{\imgSize}{1}
\plotAll{figures_first_room}{7}{\imgSize}{1}\plotAll{figures_first_room}{8}{\imgSize}{1}
\plotAll{figures_second_room}{6}{\imgSize}{1}\plotAll{figures_second_room}{7}{\imgSize}{1}
\plotAll{figures_second_room}{8}{\imgSize}{1}\plotAll{figures_third_room}{1}{\imgSize}{1}
\plotAll{figures_third_room}{2}{\imgSize}{1}\plotAll{figures_third_room}{4}{\imgSize}{1}\caption{A variety of example images from \blenderName on the SUNCG dataset \cite{suncg}}
\label{fig:example_imgs}
\end{figure}

\begin{figure}
\plotAll{figures_fourth_room}{1}{\imgSize}{0}\plotAll{figures_fourth_room}{3}{\imgSize}{0}
\plotAll{figures_fourth_room}{5}{\imgSize}{0}\plotAll{figures_fourth_room}{31}{\imgSize}{0}\end{figure}

\section{Future Work}
\label{ref:future_work}

As this project is still under heavy development, we still have major updates planned.
\begin{itemize}
\item Physics integration, where objects should get a gravity attribute to make them fall into a container
\item Object swapper, which will randomly swap objects from the same category
\item More samplers, we already offer some basic samplers for lights and cameras, but we want to extend to this to a variety of samplers
\item Decreasing render speed, by avoiding recalculation of the scene 
\end{itemize}

\bibliographystyle{IEEEtran}
\bibliography{bib}

\end{document}